# An introduction to synchronous self-learning Pareto strategy

Ahmad Mozaffari[1,*], Alireza Fathi[2]

1: Systems Design Engineering Department, University of Waterloo, ON, Canada

2: Department of Mechanical Engineering, Babol University of Technology, Babol, Iran

* Corresponding Author, E-mail: amozaffari10@yahoo.com

**Abstract**

In last decades optimization and control of complex systems that possessed various conflicted objectives simultaneously attracted an incremental interest of scientists. This is because of the vast applications of these systems in various fields of real-life engineering phenomena that are generally multi-modal, non-convex and multi criterion. Hence, many researchers utilized versatile intelligent models such as Pareto based techniques, game theory (cooperative and non-cooperative games), neuro-evolutionary systems, fuzzy logic and advanced neural networks for handling these types of problems. In this paper a novel method called Synchronous Self Learning Pareto Strategy Algorithm (SSLPSA) is presented which utilizes Evolutionary Computing (EC), Swarm Intelligence (SI) techniques and adaptive Classical Self Organizing Map (CSOM) simultaneously incorporating with a data shuffling behavior. Evolutionary Algorithms (EA) which attempt to simulate the phenomenon of natural evolution are powerful numerical optimization algorithms that reach an approximate global maximum of a complex multivariable function over a wide search space and swarm base technique can improved the intensity and the robustness in EA. CSOM is a neural network capable of learning and can improve the quality of obtained optimal Pareto front. To prove the efficient performance of proposed algorithm, authors utilized some well-known benchmark test functions. Obtained results indicate that the cited method is best suit in the case of vector optimization.

## I. INTRODUCTION

Application of soft computational systems in management, optimization and control of engineering systems is apparent. There are many intelligent models that utilize iterative stochastic/deterministic approaches, logical prediction, clustering methodologies and other advanced soft computing techniques for optimization and control of complex engineering systems [1]. Hence, many researchers focused on implementing and inspiring variety of these models by applying natural phenomena like human organisms and animals' behaviors [2]. The feedback from the new research papers indicates that soft computing techniques and machine learning methodologies attract incremental attention of scientists because of their reliability and robustness in the field of optimization and control [3]. The Pareto based algorithm that was first developed by the engineer/economist Vilfredo Pareto, is one of the most applicable techniques in vector optimization field [4]. Optimal solutions for a multi-objective problem, defined by applying Pareto's concept, are currently called as Pareto-optimal solutions. A solution is Pareto-optimal if it is non-dominated by other solution, which is superior at least in case of one objective function value, respecting to the other solutions. The Pareto optimal solutions are rather a class of solutions, forming a surface in objective function space which is generally called Pareto front [4]. Many researchers focused on this type of optimization algorithms because of its reliability in finding non-dominated optimum solutions instead of one optimum solution [3]. This feature helps the researcher to compare the obtained solutions and select a proper solution due to the real condition

In this paper a new method will be proposed and also the effective of the unsupervised machine learning in finding further non-dominated solutions with observing obtained Pareto front will be scrutinized. Obtained results evidence the robustness of proposed learnable multi-objective algorithm in finding the proper optimal front. It is also observed that, proposed method maintains the intensity and diversity of optimal front altogether.

## II. SYNCHRONOUS SELF LEARNING PARETO STRATEGY

Synchronous Self Learning Pareto Strategy Algorithm (SSLPSA) is a hybrid Pareto based algorithm which possessed some computational controlling characteristics simultaneously [15-20]. These features are: (1) synchronous parallelism, (2) unsupervised self adaptive machine learning called Kohonen map, (3) Quasi Artificial Bee Colony (QABC) communication system, (4) Tournament Based Genetic Algorithm (TBGA), (5) a fast and elitism non-dominated sorting, (6) an external archive for collecting non-dominated solutions, (7) a sharing factor ($\xi$) that controls the initial seeding of two stochastic optimizing operators i.e. QABC and TBGA in each iteration and (8) some major controlling parameters. All of the cited techniques that established the construction of proposed algorithm will be clarified precisely.

### A. Synchronous Parallelism

Parallelism is one of the most well-known techniques that makes an important contribution for improving the strength and efficiency of the modern and classical optimization programs in handling with multi-criterion and multi-modal problems [5]. Parallel programming typically divided into two main parts:

synchronous and asynchronous programming. The basic idea behind most parallel programs is accelerating the speed of optimization procedure by dividing complex tasks into finite number of subtasks which are interconnected and a sharing of information take place between them [6]. In most cases this policy will lead to more accurate solutions and improve the robustness of the optimization algorithm. In this paper a synchronous parallel strategy will be introduced in which independent elements (optimizing algorithms) are capable of shuffling the data between each other and operate independently. In order to verify the dominance of proposed method, authors developed an additional self learning Pareto strategy that fused two optimization method in a cascade form. After finite number of independent runs with different random seeding, the following listed results have been observed:

- Proposed parallel method i.e. SSLPSA possessed an acceptable quality in finding diverse batch of solutions with higher intensification comparing to a cascade Self Learning Pareto Strategy (CSLPS) model.

- Independency of two optimizing methods guaranteed the higher robustness (lower variance) of proposed algorithm in contrast with CSLPS and other well-known optimizing methods like SPEA 2 and NSGA-II.

- It has been observed that SSLPSA and CSLPS consume higher CPU time comparing to SPEA 2 and NSGA-II because of their synchronous and cascade structures respectively. This short coming can be modified by applying an asynchronous parallel algorithm; however, a synchronous model can be inspired with lower facilities and this feature justifies the application of synchronous parallel model.

- The gained results illustrate that SSLPSA has an explicit advantage in finding non-dominated solutions. This is the direct feedback of applying independent multi swarm optimizing techniques (QABC and TBGA).

*B. Machine Learning and Kohonen Map with Conscience Mechanism*

Application of hybrid evolutionary-learning algorithms had been begun by Michalski's [7] researches who hired machine learning technique and evolutionary algorithm to generate new population. These types of algorithms are simply called Learnable Evolutionary Models (LEMs). After that many researchers focused on this concept and developed new models and improvements. Ammor and Rettinger [8] applied Self Organizing Map (SOM), sometimes known as Kohonen map, to improve diversity and avoid fast convergence. SOM approximates the probability of density of input data distribution. Kobuta et al. [9] developed SOM for reproduction new seeds in GA. In this paper an adaptive SOM has been applied for detecting the features of obtained non-dominated solutions in current optimal Pareto front as on-line input data. The results show that this peculiar unsupervised network can play an important role in finding an intense front with acceptable diversification. In the following subparts the architecture of classic self organizing map with adaptive gradient learning rule and conscience mechanism will be summarized briefly.

1) *Adaptive Self Organizing Map*

Adaptive SOM which was first proposed by Kohonen is a type of unsupervised Artificial Neural Network (ANN) that automatically adapts the learning rate and neighborhood function of neuron weights independently [10]. One of its major applications is to normalize all distance calculation between any input vector (non-dominated solutions) and the neuron's weight vector by providing a suitable topological ordering for input distribution. Fig. 1 exposes a schematic of weight adaption during the process.

Generally, a SOM network with conscience mechanism uses the following learning rule:

$$W_j^{n+1}(t) = W_j^n(t) + y_j(t).h_j(n).\left(R_i^n(t) - W_j^n(t)\right), t = 1,2,\dots,T \quad (1)$$

where $t$ is sub-generation in SOM network and $n$ represents the SSLPSA generation number. $y_j(t)$ is a controlling parameter that leads weight vectors to a none dominate solution which was transferred from external archive to network as an on-line input. In other word if the input's fitness value $f_R$, which is a none-dominated solution, is lower than $f_{Wj(n)}$ then $y_j(t) = 1$ and neuron center moves toward the none-dominated solution (networks input) otherwise $y_j(t) = 0$ and neuron center does not approach the solution. This can mathematically be expressed as:

$$y_j(t) = \begin{cases} 1 \ if \ R(t) \ dominate \ Wj(t) \\ 0 \quad\quad\quad otherwise \end{cases} \quad (2)$$

$W_j^{n+1}(t)$ refers to updated weight vector and $W_j^n(t)$ is the old weight vector. $\|R_i^n - W_j^n\|$ represents the distance between input vectors where $R_i^n$ is the *i-th* non-dominated solution in *n-th* generation. The learning rate which is a descending function is defined as the following:

$$h_j(t+1) = h_j(t) + \alpha\left(f\left(\frac{1}{s_f.sl(t)}\|R_j^n(t) - W_j^n(t)\|\right)\right) \quad (3)$$

The learning rate parameter $h_j(0)$ should be initialized with value close to unity. $\alpha$ obtains any arbitrary value between 0 and 1. $s_f$ is a descending constant and should be set due to the problem condition. In this paper we set $s_f = 1000$. Function $f(.)$ should be designed in a manner that following criteria is satisfied:

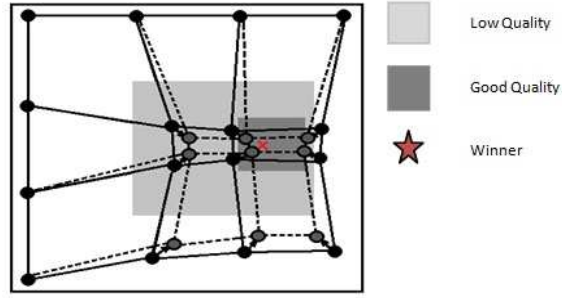

**Figure 1.** Weight adaption during the process

$f(0) = 0$, $0 \leq f(z) \leq 1$, and $\frac{df(z)}{dz} \geq 0$ for positive values of z.

In this paper $f(z)$ set as:

$$f(z) = 1 - 1/1 + z \qquad (4)$$

Shah-Hosseini and Safabakhsh [10] produced a scaling value for a 2-D input. In this paper scaling value is extended to a n-D input due to the number of our decision parameters and our problem conditions. Scaling value *sl* adjusts using following equation:

$$sl(t+1) = \sqrt{(\sum_{i=1}^{n} E_k^i(t+1)^{n-i}(-1)^{i+1})^+}, k = 1 \qquad (5)$$

$$E_k^i(t+1) = E_k^i(t) + \mu_i \left( R_k^i(t) - E_k^i(t) \right) \qquad (6)$$

where *i* represents the number of variables in each solution. $E_k^i(0)$ initialized with some small random values.

*2) Conscience mechanism*

Conscience mechanism is applied in order to revive the dead units (weights) in neuron center. The scheme of this procedure is depicted in Fig. 2. Dead unit is a term that refers to weights with a trivial chance of learning and adaption during the progress. The policy of repairing these units is often called conscience mechanism. In this paper a simple well-known mechanism is utilized which tunes the bios of each node (neuron) by the following formula [11]:

$$b_i(t+1) = \begin{cases} 0.8\, b_i(t) \\ \text{or} \\ b_i(t) - 0.3 \end{cases} \qquad (7)$$

## C. Quasi Artificial Bee Colony Communication System

Classical Artificial Bee Colony (ABC) which first developed by Karaboga and Basturk [12] is a powerful optimizing method that can be categorized in Swarm Intelligent (SI) techniques. Generally, the colony of artificial bees contains three groups of independent active bees: employed bees, onlooker bees and scouts. Half of the colony includes the employed bees and the remaining bees are onlookers. Each employed bee focuses on one source of food so the food number is equal to half of population (number of employed bees). These bees try to improve they source quality each time but if they fail after a finite number of trails, they abandon the food source and become a scout bee. Karaboga and Basturk [12] prove that scout bees have an important contribution in the successful performance of ABC since they lead the colony to new solution areas. In this paper, authors proposed a Quasi Artificial Bee Colony (QABC) that omitted the scout bee phase. This will lead to an operator that completely concentrates on local areas and reduce the amplitude of agents shaking in solution area. This policy controls the rate of diversity and forbids the external mutation in SSLPSA since the diversity has been guaranteed in the mutation phase of TBGA. The authors find the following advantages in utilizing QABC instead of ABC:

- QABC enhanced the intensity of candidate solutions and damps the additional shaking that takes place in ABC. The results expose that applying the scout bee phase (in ABC) and mutation phase (in TBGA) altogether leads to an optimizing method that automatically distributes the optimizing agents and forbids the exploring (intensification) capability and this leads to weak results.

- Scout bee phase can descend the chance of obtaining optimal Pareto front in SSLPSA since it omits all of the superior and inferior solutions trapped in local areas (Haming valley).

- Applying QABC which is an independent optimizing model in SSLPSA will improve the robustness of final Pareto front, since the onlooker phase concentrates on neighbor searching without considering the interactions that take place in TBGA.

- It can be interpreted that QABC is a simple algorithm that does not consume so much CPU time and also plays an efficient role as a co-evolutionary optimizing method.

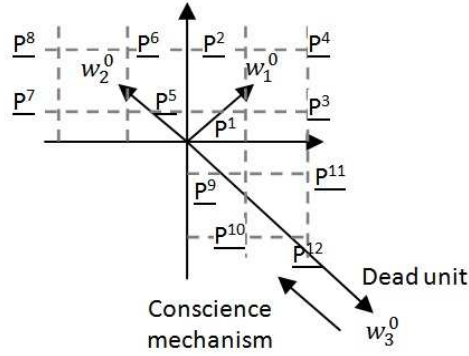

**Figure 2.**   Schematic of Conscience Mechanism

### D. Tournament Based Genetic Algorithm

Tournament Based Genetic Algorithm (TBGA) is a well-known EA in which the selection mechanism occurs by exerting a tournament between candidate challenger chromosomes [3]. This technique is best suit in the case of constraint optimization since it obeys the following rules:

- The solution which possesses higher diversity or belongs to non-dominated Pareto front will be selected through tournament selection.

- Among the solutions that exceed from constraint boundaries, the solution with lower deviation will be selected.

In this paper authors utilized this type of GA for granting the above rules. *Fig. 3* indicates the scheme of tournament selection mechanism.

### E. Fast and Elitism Non Dominated Sorting

The idea of fast and elitism non-dominated sorting which was first developed by Deb *et al.* [13], concentrates on reducing the computational complexity of prior types of non-dominated sorting methods. In this model the solutions that belong to first non-dominated front have their domination count as zero ($n_p = 0$). First we assume that all solutions belong to first Pareto front with $n_p = 0$. In the next step, we compare each solution *p* with other solutions *q*. If candidate solution *p* dominates other solutions it will be saved in the first Pareto front with $n_p = 0$ and the dominated solutions will be added to the set of solutions that are dominated by *p* ($S_p$) and their domination count is reduced by one. After that, a non-dominated sorting algorithm will be applied for the solutions that are listed in $S_p$. The new obtained non-dominated solutions will be saved in Q as the members of second front. This procedure continues till all solutions are categorized in their appropriate fronts. Figure 4 indicates the schematic of fast and elitism non-dominated sorting in a two dimensional objective space.

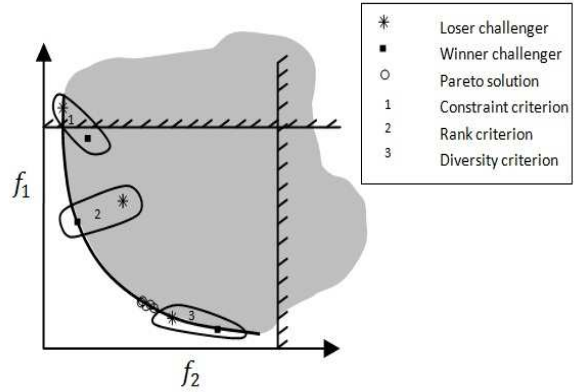

**Figure 3.**      Tournament selection mechanism

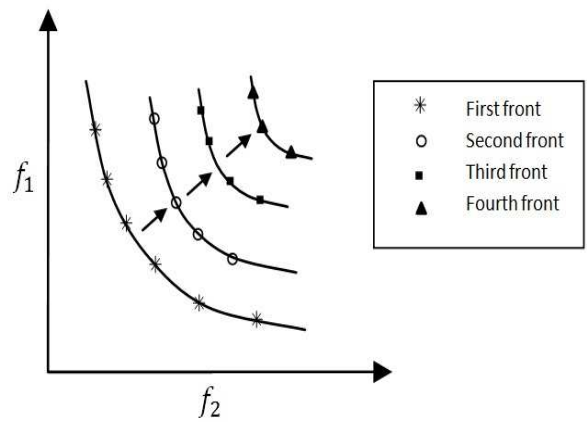

**Figure 4.**      Non-dominated sorting procedure

## F. *External Archive*

In SSLPSA an external archive has been utilized to save the global non-dominated solutions. This technique reduces the computational complexity when the algorithm derives to high number of non-dominated solutions since its task is just to retain the obtained non-dominated solutions in each generation. Besides, defining an external archive provides some elitism by preserving non-dominated solutions. The feedback of applying this policy illustrates that the speed of algorithm does not decline considering the advantage of higher number of obtained non-dominated solutions comparing to NSGA-II.

## G. *Sharing Factor*

Sharing factor (ꜱ) is a controlling parameter that determines the amount of solutions that should be seeded for the optimizing algorithm. Sharing factor spans in 0 and 1 and due to the optimizing problem it can be constant or adapts itself during the optimizing process. Following feedbacks can be listed for application of solution sharing among the optimizing operators:

(1) By shuffling data between optimizing operators, the speed of SSLPSA can be controlled. Since TBGA consumes more CPU time comparing to QABC, in optimizing normal problems we can thrust a higher amount of candidate solutions in QABC phase in order to enhance the speed of SSLPSA while in optimizing a rigid multi-modal problem, the results indicate that it is an acceptable policy to flux a higher percent of solutions in TBGA.

(2) As a view of control, by defining a sharing factor, we produced an on-line multi-player optimizing methodology in which each of these two operators can be conveniently omitted from optimizing cycle. Besides, for each optimizing problem, the effect of these two operators in deriving to optimum front can be arithmetically verified.

(3) By considering the above advantages, it can be interpreted that sharing factor is completely useful for controlling a multi-swarm optimizing method.

The control diagram of a multi operator optimizing model with a central actuator is shown in *Fig. 5*.

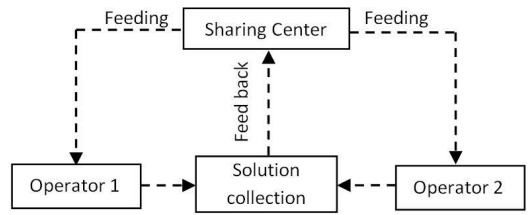

**Figure 5.**   Controlling diagram of multi-swarm optimizing operator

TABLE I.   SSLPSA CONTROL PARAMETERS

| CP | $P_{mut}$ | $P_s$ | $\xi$ | $n$ | $t$ | $s_f$ | $\alpha$ | $\mu$ |
|---|---|---|---|---|---|---|---|---|
|  | 0.1 | 30 | 0.65 | 100 | 5 | 1000 | 0.4 | 0.5 |

## H. SSLPSA Controlling Parameters

Basically, SSLPSA possessed a limited number of controlling parameters (CP) which are expressed in Table 1.

## I. The Structure of SSLPSA

In this section, pseudo code of proposed method will be given:

- *Step 0*: Define algorithm initial parameters such as mutation probability ($P_{mut}$), number of neurons ($Nu_{TBGA}$ and $Nu_{QABC}$) in SOM center for each phase, sharing factor ($\xi$), pool size, number of generation, population size ($P_s$), descending constant ($s_f$), $\alpha$, $\mu$ and stopping criterion. Set $n = 1$ and start the process.

- *Step 1*: Randomly generate $P_s$ solution for the initial population $P_1$.

- *Step 2*: Share (shuffle) the solutions into QABC phase and TBGA phase due to the sharing factor ($\xi$). $\xi$ is a random number from a uniform distribution. Lead $(1 - \xi) * P_s$ of solutions in QABC operator phase ($P_{QABC}$) and the rest of them in genetic operator ($P_{TBGA}$).

- *Step 3*: Evaluate the fitness of QABC solutions (foods) in $P_{QABC}$ and rank them based on non-dominated sorting and crowding distance.

- *Step 4*: Define random weight vectors for SOM unit center in QABC operator phase ($Nu_{QABC}$) in a uniform stochastic distribution manner spanning to problem solution space. Evaluate the fitness of weight vectors.

- *Step 5*: Train the weight vectors in SOM center ($W_j^n, j = 1, 2, \ldots, Nu_{QABC}$) using obtained non-dominate solutions (elite bees) in the current $n_{th}$ generation.

- *Step 6*: Generate new weight vector $W_j^{n+1}$.

- *Step 7*: if the new weights dominated old ones, replace old ones with new ones. In other words move the SOM mobile units toward better area. If they do not dominate each other, save the new non-dominate weights in external archive.

- *Step 8*: Apply the employed bees for neighbor search (as agents that perform near the $P_{QABC}$)

- *Step 9*: Evaluate the fitness of new obtained solutions.

- *Step 10*: Sort the new solutions based on non-dominated sorting and crowding distance to evaluate their fitness.

- *Step 11*: Select a food source (solution) and employed the onlooker bees in order to perform a neighbor search near the chosen solution and a greedy selection based on the evaluated fitness.

- *Step 12*: if all of the onlooker agents contribute in searching go to the next step, otherwise return to step 11.

- *Step 13*: Export the obtained solutions in QABC phase to the collection site.

- *Step 14*: Evaluate the fitness of TBGA solutions (chromosomes) in $P_{TBGA}$ and rank them based on none-dominated sorting and crowding distance.

- *Step 15*: Perform a same Treat for SOM center in TBGA phase. In other words regard $Nu_{TBGA}$ instead of $Nu_{QABC}$ and repeat steps 4 to 7 respectively.

- *Step 16*: Generate a random number with uniform distribution. If the random number is less than $P_{mutation}$ produce children using mutation operator, else produce children using crossover due to the pool size.

- *Step 17*: Evaluate the fitness of produced solutions and combine them with old population. Rank all of the solutions using non-dominate sorting and crowding distance.

- *Step 18*: Select $\xi * P_s$ best solutions from current population.

- *Step 19*: Export the obtained solutions in TBGA phase to the collection site.

- *Step 20*: if the stopping criterion is satisfied, go to step 21, otherwise go to step 3.

- *Step 21*: latest population, the weight vectors in both SOM centers and also the recorded solutions (Archived ones) are considered as the final solution.

- *Step 22*: Stop

*Fig. 6* shows the block diagram of the SSLPSA algorithm.

### III. RESULTS AND DISCUSSION

#### A. Defining the Comparative Metrics and Test Algorithm's Controlling Parameters

For validating the performance of proposed learnable optimizing method, the results have been compared to some of well-known multi-objective optimizing models such as NSGA-II, SPEA 2, INSGA and Multi Objective Bee Algorithm (MOBA) that proposed by Pham and Ghanbarzadeh [2]. Table 1 indicates the controlling parameters for SSLPSA. For running NSGA-II it has been assumed that the mutation probability ($P_{mut}$) is 0.1, number of chromosomes are 100, the pool size is 20, the number of challengers in selection tournament are 2, number of iterations are 100. SPEA 2 used all parameters set in NSGA-II with an additional archive size of 200. In this paper MOBA applied with hive size of 80, 5 elite patches in each of iterations ($e = 5$) and 9 best sights ($m = 9$) in which 11 bees recruited around elite patches and 7 bees independently search around best sights. A same INSGA that proposed in [14] is used for further validation. The stopping criterion is assumed to be 100 generations for all of the multi-objective optimizing models.

Seven well-known dual-objective benchmarks nominally ZDT1, ZDT2, ZDT3, ZDT4, ZDT6, SCH and FON are extracted from [3] to verify the performance of SSLPSA.

Some metrics are applied to check the robustness and efficiency of SSLPSA in multiple runs (30 different runs). These metrics are extracted from [3]. The first metric ($\gamma$) measures the convergence of obtained Pareto front to a detailed sample of the true one. The second metric ($\Delta$), called diversity metric, illustrates the extension of non-dominated solutions over the obtained Pareto front. The third one is Inverted Generational Distance (IGD) which calculates the Euclidian distance of each solution in obtained Pareto front with respective solutions in those of true one. The last one is SPREAD that indicates the spread of obtained solutions in the Pareto front. It is important to note that SPREAD and diversity act near similar. However, applying both of them yields more reliable feedback on the performance of an optimization algorithm. The definitions of these metrics are completely explained in Refs [3, 14].

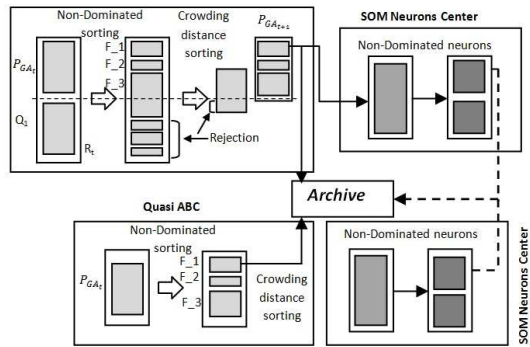

**Figure 6.** Block diagram of SSLPSA

## B. Unconstraint Optimization

In this section, the algorithm has been tested with eight test problems. Then the cited comparative metrics are used for comparing the efficiency of SSLPSA with NSGA-II, SPEA 2 and MOBA. *Fig.* 7 ($a_1$) to ($g_3$) indicates the best obtained Pareto front for each of unconstraint tested problems in 30 runs.

*Fig.* 7 ($a_1$) to ($a_3$) indicates the obtained Pareto for ZDT1 using SSLPSA, MOBA, NSGA-II and INSGA. It can be seen that SSLPSA, NSGA-II and INSGA approximate the true Pareto front. However, MOBA shows weak results in all 30 runs. One of the important advantages of SSLPSA is finding a high amount of Non-dominated solutions (near 3500 solutions) in an acceptable run time for ZDT1. This refers to the high intensification quality of SSLPSA. The results also illustrate that proposed method has a high robustness in converging to optimal Pareto front comparing to other well-known methods. The efficient Pareto fronts in ZDT2, ZDT4, ZDT6 and SCH are also easily obtained using SSLPSA while MOBA converge to local Pareto fronts. Only in FON problem proposed method was not successful to find the whole Pareto front. For further authenticity in the performance of SSLPSA the defined contrasting metrics are tabulated in Tables 2 and 3. It can be seen that SSLPSA is more capable to converge to optimal Pareto front comparing to NSGA-II. However NSGA-II shows better result for ZDT3 and FON. This is because of the SSLPSA local convergence. Consequently it can be concluded that SSLPSA possessed better quality in converging to true Pareto front in contrast to NSGA-II. The gained results for $\Delta$ and SPREAD also suggest that the proposed method covers the whole Pareto front in most test functions. It seems that SSLPSA is just failed to find the complete Pareto front for FON problem.

## IV. CONCLUSIONS

In this study, a novel self learning optimizing method was developed that utilize the evolutionary concepts and machine learning procedure altogether. For evaluating the performance of the proposed method it was tested with different benchmark problems and compared with some well-known optimizing methods nominally NSGA-II, INSGA, SPEA 2 and MOBA. Also some comparing metrics such as diversity, convergence, spread and inverted generational distance were developed for deriving to some reliable conclusions. Besides, an additional metric that is the number of obtained non-dominated solutions was utilized for verifying the capability of optimizing methods in finding a variety of potential solutions. Obtained results evidence the explicit potential of proposed method in both robustness and quality. In most cases SSLPSA outperformed other optimizing methods. The gained results also indicate the high potential of hybrid evolutionary decision support machines in the field of vector optimization. Further investigations lies on the field of evolutionary games. In the next works the concept of non cooperative games and decision support machines will be fused to a differential evolutionary optimizing method.

TABLE II. QUALITY MEASURES OF NSGA-II AND SSLPSA

| Problem | Mean value of IGD | | Mean value of Δ | | Mean value of γ | | Mean value of SPREAD | |
|---|---|---|---|---|---|---|---|---|
| | NSGA-II | SSLPSA | NSGA-II | SSLPSA | NSGA-II | SSLPSA | NSGA-II | SSLPSA |
| ZDT1 | 1.91E-04 | **1.51E-04** | 4.04E-01 | **1.7E-01** | 1.82E-04 | **1.58E-04** | 3.83E-01 | **1.56E-01** |
| ZDT2 | 1.88E-04 | **1.28E-04** | 3.94E-01 | **1.84E-01** | 1.69E-04 | **1.31E-04** | 3.52E-01 | **1.69E-01** |
| ZDT3 | **2.59E-04** | 2.64E-04 | **4.64E-01** | 6.87E-01 | **2.59E-04** | 2.64E-04 | 7.49E-01 | **6.76E-01** |
| ZDT4 | 1.84E-04 | **1.71E-04** | 3.84E-01 | **1.44E-01** | 1.84E-04 | **1.71E-04** | 3.96E-01 | **1.23E-01** |
| ZDT6 | 1.59E-04 | **1.43E-04** | 5.20E-01 | **1.52E-01** | 1.69E-04 | **1.55E-04** | 4.80E-01 | **1.40E-01** |
| SCH | 2.04E-04 | **1.84E-04** | 6.10E-01 | **2.45E-01** | 2.14E-04 | **1.74E-04** | 6.46E-01 | **2.32E-01** |
| FON | **1.95E-04** | 3.05E-04 | **3.83E-01** | 7.62E-01 | **1.99E-04** | 2.87E-04 | **3.94E-01** | 7.46E-01 |

\* The data with bold fonts represent better solutions found using the two algorithms

TABLE III. ROBUSTNESS MEASURES OF NSGA-II AND SSLPSA

| Problem | Standard deviation of IGD | | Standard deviation of Δ | | Standard deviation of γ | | Standard deviation of SPREAD | |
|---|---|---|---|---|---|---|---|---|
| | NSGA-II | SSLPSA | NSGA-II | SSLPSA | NSGA-II | SSLPSA | NSGA-II | SSLPSA |
| ZDT1 | 1.08E-05 | **6.82E-06** | 1.79E-02 | **8.06E-04** | 0.08E-05 | **6.82E-06** | 3.14E-02 | **8.06E-04** |
| ZDT2 | 8.36E-06 | **1.28E-08** | 3.14E-02 | **9.69E-03** | 2.36E-06 | **1.28E-08** | 7.25E-02 | **9.69E-03** |
| ZDT3 | 1.16E-05 | **2.17E-08** | 3.56E-02 | **6.11E-03** | 4.11E-05 | **2.17E-08** | 1.49E-02 | **6.11E-03** |
| ZDT4 | 9.86E-06 | **8.26E-07** | 3.24E-02 | **1.43E-02** | 4.86E-06 | **8.26E-07** | 2.94E-02 | **1.43E-02** |
| ZDT6 | 1.24E-05 | **8.22E-06** | 2.44E-02 | **9.49E-04** | 9.94E-06 | **8.22E-06** | 4.49E-02 | **9.49E-04** |
| SCH | 2.13E-05 | **1.84E-05** | 2.66E-02 | **2.02E-02** | 1.33E-05 | **1.84E-06** | 6.46E-02 | **2.02E-02** |
| FON | 2.95E-05 | **3.05E-06** | 2.63E-02 | **8.12E-03** | 4.55E-05 | **3.05E-06** | 3.94E-02 | **8.12E-03** |

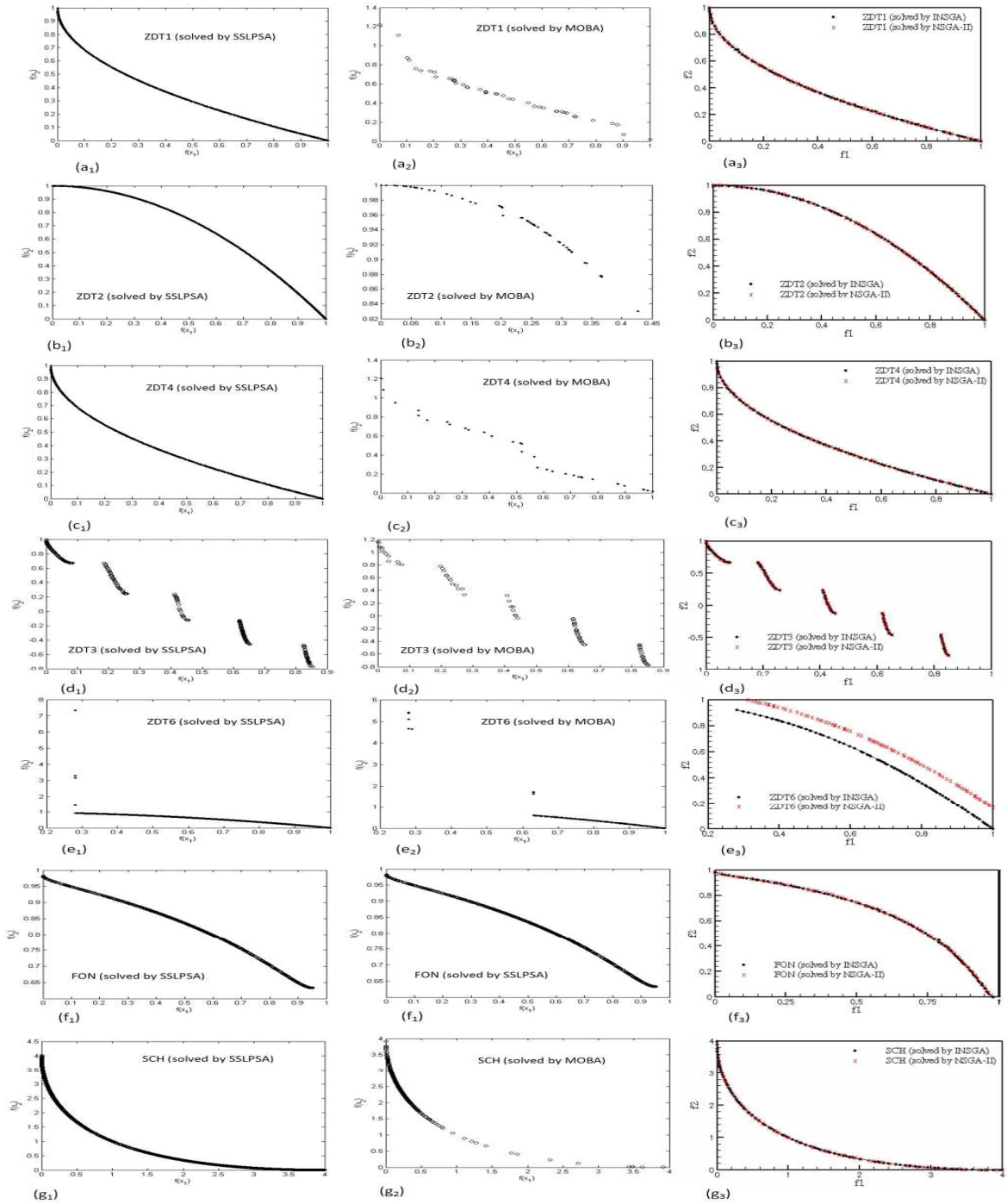

**Figure 7.** Comparison of the obtained results using SSLPSA, SPEA 2, NSGA-II and INSGA for problems (a) ZDT1, (b) ZDT2, (c) ZDT4, (d) ZDT3, (e) ZDT6, (f) FON, (g) SCH